# Parallel 3DPIFCM Algorithm for Noisy Brain MRI Images

*Arie Agranonik[(a)], Maya Herman[(b)], Mark Last[(c)]*

(a) arieag@gmail.com, Open University, Israel.
(b) maya@openu.ac.il, Open University, Israel.
(c) mlast@bgu.ac.il, Ben Gurion University of the Negev, Israel.

## Abstract

In this paper we implemented the algorithm we developed in [1] called 3DPIFCM in a parallel environment by using CUDA on a GPU. In our previous work we introduced 3DPIFCM which performs segmentation of images in noisy conditions and uses particle swarm optimization for finding the optimal algorithm parameters to account for noise. This algorithm achieved state of the art segmentation accuracy when compared to FCM (Fuzzy C-Means), IFCMPSO (Improved Fuzzy C-Means with Particle Swarm Optimization), GAIFCM (Genetic Algorithm Improved Fuzzy C-Means) on noisy MRI images of an adult Brain.

When using a genetic algorithm or PSO (Particle Swarm Optimization) on a single machine for optimization we witnessed long execution times for practical clinical usage. Therefore, in the current paper our goal was to speed up the execution of 3DPIFCM by taking out parts of the algorithm and executing them as kernels on a GPU. The algorithm was implemented using the CUDA [13] framework from NVIDIA and experiments where performed on a server containing 64GB RAM , 8 cores and a TITAN X GPU with 3072 SP cores and 12GB of GPU memory. We ran the experiments for each algorithm using python with scientific computing libraries like numba and numpy that have near native performance for matrix calculations both on GPU and CPUs respectively.

In the experiments our main metric was execution time in seconds. We conducted experiments with the parallel version of the algorithm to itself in CPU version and also to IFCMPSO, GAIFCM which are genetic algorithm counterparts of the 3DPIFCM algorithm. We also perform comparison to the original FCM algorithm which performs the fastest but lacks any optimization that counters for noise added to images as produced in CT or MRI scans. In our experiments we used synthetic data of varied image sizes between 32x32 pixels to 854x854 pixels. We also ran an experiment on a standard Brainweb Volume to examine the speedup in real world conditions.

Our results show that the parallel version of the algorithm performs up to 27x faster than the original sequential version and 68x faster than GAIFCM algorithm. We show that the speedup of the parallel version increases as we increase the size of the image due to better utilization of cores in the GPU. Also, we show a speedup of up to 5x in our Brainweb experiment compared to other generic variants such as IFCMPSO and GAIFCM.

Those results prove that 3DPIFCM runs very well on a parallel execution environment much like the original FCM. Also, we show that this algorithm can be an adequate replacement to FCM in noisy image environments under time limited medical conditions. Since the algorithms is generic and doesn't depend on a training set there is a potential to do clustering of different modalities that exist in industry such as CT, MRI and Ultrasound. All those modalities use 3D imaging and have a noisy component that can be cleared by our algorithm during segmentation. The broader potential of our work is the ability to introduce the algorithm into different clinical settings by presenting a competitive execution runtime and giving high quality segmentation results.

## 1. Introduction

The original FCM[5] algorithm uses feature attraction to segment each pixel to their correct cluster. FCM implementation are normally very fast when clustering a single image and take less than a second to execute (see section 5). Nevertheless FCM fails to cluster correctly when encountering noisy images. To counter this problem Shen [16] developed IFCM which is using feature attraction and neighborhood attraction parameters in the distance function of the algorithm. The immediate results of this approach is much better segmentation in noisy images. However, this algorithm is expensive and had some issues with the cost of optimization of its parameters λ, ξ. The algorithm used a neural network for optimization which was costly and reached local minima at some points. This approach is supervised and requires to first train the network. Also, the algorithm can take many seconds to execute as images grow in size.

To tackle the neural network training problem some genetic optimization algorithms such as GAIFCM [8] and IFCMPSO [7] where used with greater success. They simplified the optimization procedure of IFCM but still only used 2D features from the neighborhood of the target pixel. Also, since neighborhood attraction was used the runtime of this algorithm increased by the factor of the neighbors in each pixel's vicinity. As we show in section 5 the runtime of those IFCM variants can reach hundreds of seconds depending on the size of the image and initial parameters.

To speed up the execution of FCM and make it parallel many works were conducted to enable the implementation of the FCM algorithm on a GPU [3][19][20][2][11][18][15][14]. Those works use different modifications to FCM to enable fast GPU implementations. However, as of writing this paper we couldn't find any IFCM or genetic IFCM variants that run on a GPU and hence implemented 3DPIFCM. We give a brief state of related work for FCM based implementations here.

Li et al [11] implemented a parallel FCM by modifying the calculations of the membership function to run on a GPU using CUDA. The achieved a 10x speedup to the original algorithm running on a CPU. They used natural images of sizes 53k-101k on an NVIDIA GTX 260. Mahmoud et al. [2] implemented an FCM variant called brFCM and showed that it was 23x faster than Soroosh [18]. The modified algorithm uses data reduction and aggregation to cluster image data. The algorithm was tested on medical MRI and CT images of sizes 350x350 and 512x512.

In the last years GPUs became a general purpose processing units for parallel algorithms because of new frameworks and methods [13]. Since the release of CUDA framework in 2007 by NVIDIA the GPU computation became more accessible to non-graphics algorithms and applications. Parallel algorithms using GPU have been reported to perform up to 245 times the speeds of a single CPU based algorithm [3].

Shalom et al. [15] implemented a parallel FCM algorithm on two NVIDIA GeForce 8500 cards and reached 73 fold speedup. Another experiment on 65k yeast gene expression data set of 79 dimension yielded a 140x speedup for this algorithm. Rowinska and Goclawski [14] showed a parallel FCM algorithm on polyurethane foam with fungus color images and compared to sequential FCM using C++ and MATLAB. Using C++ they reached a 10x speedup on 310k pixels and 50-100x speedup on the MATLAB version for 260k pixels.

Mishal et al [3] used a reduction technique to aggregate the membership and cluster vectors. They compared to sequential version of FCM and reached 245x speedup using shared GPU memory on a NVIDIA Tesla C2050 GPU. For the implementation of their algorithm they used C language and tested on brain images segmenting WM, GM, CSF segments.

In this paper we look at the bottlenecks of the new 3DPIFCM algorithm and design the entire algorithm on a parallel environment using the CUDA framework. The 3DPIFCM is a 3D algorithm in nature. It's performing the optimization of clearing noise by utilizing the 3D environment of each voxel it's trying to segment to its respective cluster. We recognize that the majority of the work can be parallelized by sending the computation of each voxel clustering to the GPU. In such a way we utilize the massive parallel execution environment of CUDA and send small independent computations to the GPU per voxel.

Our contribution in this paper is rewriting the original 3DPIFCM algorithm that was developed in [1] in a parallel environment namely CUDA on a GPU. We achieve a speedup of up to 68x than the slowest 2D variant when comparing raw execution speed in the same conditions. This leads to the opportunity of our research to be implemented in clinical settings where timely results are of the essence.

The rest of this paper is organized as follows: section 2 is a brief background of 3DPIFCM. Section 3 an introduction to GPU computing and the CUDA framework. Section 4 is the parallel implementation of 3DPIFCM. Section 5 is the tooling used for implementation. Section 6 is results and comparison to CPU counterparts. Section 7 is limitations of our study and section 8 is conclusion and section 9 are further steps.

## 2. Background

3DPIFCM [1] is based on FCM[5] with a fuzzy membership matrix whereby each pixel in the image has a probability of belonging to a certain cluster. The algorithm is iterative and has a cost function (1) shown below. The cost function is the sum of all voxels and all clusters multiplied by a membership matrix and a distance function $d^2(x_i, c_j)$. The distance function in 3DPIFCM looks at the feature attraction and neighborhood attraction parameters of each voxel to the target voxel. The algorithm iterates between formulas (2) and (3) by optimizing U - membership matrix and C – cluster centers.

$$J_m(U, \bar{c}) = \sum_{i=1}^{N} \sum_{j=1}^{C} u_{ij}^m d^2(x_i, c_j) \tag{1}$$

$$u_{ij} = \frac{1}{\sum_{k=1}^{C} \left(\frac{d(x_i, c_j)}{d(x_i, c_k)}\right)^{\frac{2}{m-1}}} \tag{2}$$

$$c_j = \frac{\sum_{i=1}^{N} u_{ij}^m \cdot x_i}{\sum_{i=1}^{N} u_{ij}^m} \tag{3}$$

The cost function $d^2(x_i, c_j)$ looks at each voxel's neighborhood and feature attraction parameters by optimizing $\lambda$ and $\xi$ as shown in (4)

$$d^2(x_i, c_j) = \|x_i - c_j\|^2 (1 - \lambda H_{ij} - \xi F_{ij}) \tag{4}$$

Those parameters represent both feature attraction like in FCM and neighborhood attraction to counter noise pixels that might be interfering with the segmentation. $H_{ij}$ represents feature attraction as shown in (5).

$$H_{ij} = \sum_{r=1}^{V} W_r \frac{\sum_{k=1}^{S_r} u_{kj} g_{ik}}{\sum_{k=1}^{S_r} g_{ik}} \tag{5}$$

The feature attraction part is in $g_{ik}$ where voxel differences are calculated between $x_i$ and $x_k$ as shown in equation (6).

$$g_{ik} = |x_i - x_k| \qquad (6)$$

For neighbourhood attraction equation (7) is used. $q_{ik}^2$ represents the distance of the target voxel k from the neighbouring i voxel.

$$F_{ij} = \sum_{r=1}^{V} W_r \frac{\sum_{k=1}^{S_r} u_{kj}^2 q_{ik}^2}{\sum_{k=1}^{S_r} q_{ik}^2} \qquad (7)$$

In equation (8) the distance is used as a measure of which cluster the target voxel should belong to according to its neighbours.

$$q_{ik} = \left(X_{x_i} - X_{x_k}\right)^2 + \left(Y_{x_i} - Y_{x_k}\right)^2 + \left(Z_{x_i} - Z_{x_k}\right)^2 \qquad (8)$$

where $x_i = \left(X_{x_i}, Y_{x_i}\right), x_k = (X_{x_k}, Y_{x_k})$ and the neighborhood of $x_i$ is in equation (9). L represents a neighbourhood matrix (see [16]).

$$NB_{x_i} = \left\{ x_k \in I : 0 < \left(\left(X_{x_i} - X_{x_k}\right)^2 + \left(Y_{x_i} - Y_{x_k}\right)^2 + \left(Z_{x_i} - Z_{x_k}\right)^2 < 2^{L-1}\right)\right\} \qquad (9)$$

In 3DPIFCM the original formulas of IFCM were modified and introduced (5, 7, 8, 9 and 10). Those equations introduce a 3rd dimension to the 3DPIFCM algorithm by changing the cost function (1) to look in the 3D environment and correct for noise. Feature attraction and neighborhood attraction parameters are also changed (5, 7) so that they account for a new exponential decay parameter $W_r$ which are introduced in equation (10).

$$W_i = \frac{e^{-\frac{i}{h}}}{\sum_{r=1}^{V} e^{-\frac{r}{h}}} \text{ where } i = 1..v \qquad (10)$$

In equation (10) v and h parameters were introduced. v account for distance of the furthest voxel from the center (i.e. 3D neighborhood attraction) and h accounts for exponential decay of importance of voxels from the center to the final clustering. Optimal values for those parameters were shown for different noise levels in [1]. The final results of SIMIFCM show improvement over both FCM, GAIFCM, and IFCMPSO algorithms both in synthetic images and Brainweb T1 images.

The original 3DPIFCM is shown in algorithm 1.

|  | Algorithm 1: 3DPIFCM |
|---|---|
|  | Input: img3d - a 3D matrix of pixel intensities, c – number of clusters, v – depth parameter, h – exponential decay, z – which slice in z axis to segment, m – fuzziness, $\epsilon$ – stop criteria |
|  | Output: centers, U – membershipMatrix |
| 1. | img = img3d[z] |
| 2. | cluster_centers1, U1 = Modified_FCM (img, c, $\epsilon$, m) |
| 3. | Generate a random swarm of P particles in 2 dimensional space (we use D=2 since there are two optimization parameters). |
| 4. | Evaluate fitness of each particle in the swarm $f(\bar{x}_i(t))$ with respect to the cost function $J_m$. |
| 5. | If $f(\bar{x}_i(t)) < pbest_i$ then $pbest_i = f(\bar{x}_i(t))$ and $\bar{x}_{pbest_i} = \bar{x}_i(t)$ where $pbest_i$ is the current best fitness achieved by the i-th particle and $\bar{x}_{pbest_i}$ is the corresponding coordinate. |
| 6. | If $f(\bar{x}_i(t)) < lbest_i$ than $lbest_i = f(\bar{x}_i(t))$, where $lbest_i$ is the best fitness over the topological neighbors. |
| 7. | Change the velocity $v_i$ of each particle: $\bar{v}_i(t) = \bar{v}_i(t-1) + p_1\left(\bar{x}_{pbest_i} - \bar{x}_i(t)\right) + p_2\left(\bar{x}_{lbest_i} - \bar{x}_i(t)\right)$ $p_1$ and $p_2$ are random constants between 0 and 1. |
| 8. | Fly each particle to its new position $\bar{x}_i(t) + \bar{v}_i(t)$ |
| 9. | Go to step 4 until convergence (i.e small changes to J cost function). |
| 10. |  |
| 11. | cluster_centers2, U2, $\lambda$, $\xi = \bar{x}_{pbest_i}$ variables. cluster_centers, U= IFCM (img , $\lambda$, $\xi$. cluster_centers2, U2, v, m, c, $\epsilon$) |
| 12 | Return cluster_centers, U |

Algorithm description:
1. The z slice is assigned to a new variable
2. The modified FCM with Gaussian mixture model is run instead of random initialization of centers.

3. Generate random particles to evaluate parameters λ, ξ.
4. Steps 4-9 are running the PSO algorithm and executing the step function of 3DPIFCM at each step. The step function includes evaluating equation 1 with modified formulas 6, 8, 9
10. Take the best particle in the swarm after the swarm finished executing and get values λ, ξ, cluster centers and membership matrix.
11. Execute standard IFCM with correct λ, ξ. parameters and correct membership matrix and cluster centers.

As was shown in [1] the 3DPIFCM algorithm outperforms state of the art IFCM variants such as GAIFCM and IFCMPSO by a margin of 5-50% for synthetic data and 1-28% for simulated brain data. However, since all derivatives of FCM perform expensive optimization of $\lambda, \xi$ they run for minutes at a time when executed on a single CPU core. 3DPIFCM suffers from the same performance bottlenecks as his counterparts. Each segmentation of a single image can be very time consuming because apart from running against all voxels in the target image the algorithm needs to look at neighboring voxels per iteration. In this paper we introduce a new implementation of 3DPIFCM that reduces time considerably.

## 3. GPU computing

The standard NVIDIA GPU is built by having many processing units called scalar processors (SPs). Each SP is assigned to a block of processors which have shared block memory. This memory is very fast and is accessible to each thread running within a block. In addition, the GPU has a device memory called global memory. The modern GPU's have device memory that ranges in Gigabytes. Normally, a general purpose algorithm running on a GPU will execute thousands of threads. Each thread will have access to local (register based), shared and global memory. Each algorithm designed for the GPU must have thread contention and locks to prevent other threads modifying memory locations that are not desired.

As can be seen in Figure 1 the host on the left is the CPU and main memory. At the first stage of each parallel algorithm data is moved from host to global memory to be processed by kernels. Kernel is a small execution function used by the GPU by many threads running concurrently. Each thread that picks up data from global memory executes the kernel on this data running inside a block. The blocks are located inside the grid. Each thread may use register memory for local variables. This memory is very fast but limited in size. Shared memory is used by multiple threads running inside a block thereby utilizing sharing between local block threads for some memory operations. Figure 1 shows that general GPU structure.

Figure 1: GPU structure.

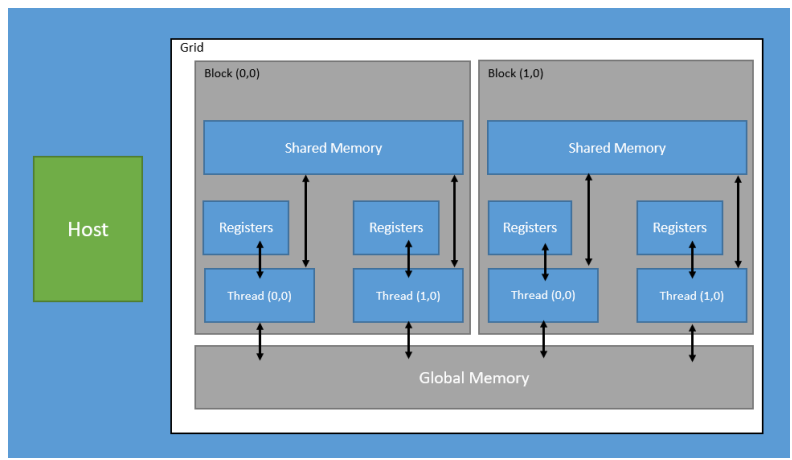

The CUDA API enables to use a new programmatic model on a GPU. This API hides hardware details from the programmer and enables compiler directives which are annotations in the code. Those directives will execute in parallel when compiled with the nvcc compiler and run directly on the GPU device. As a result, many applications and algorithms that were previously targeted to CPU's are now running in a parallel framework on a GPU [9]. Those include scientific, analytic and engineering applications [20][17].

As can be seen in figure 2 the host sends the kernel code to the GPU to be distributed in many blocks and threads on a single grid. Once the code is run each kernel will handle a different part of the data stream. The kernel will receive block id, thread id and block dimension as part of its input. As a result, it is possible to control which part of the global data the thread is going to handle by multiplying block id by block dimension and adding thread id to find a location in a global array. Because of this parallel structure it's possible to send large amounts of data to the global memory once and have the kernels process it in parallel. Figure 2 shows the CUDA processing model.

Figure 2: CUDA processing model.

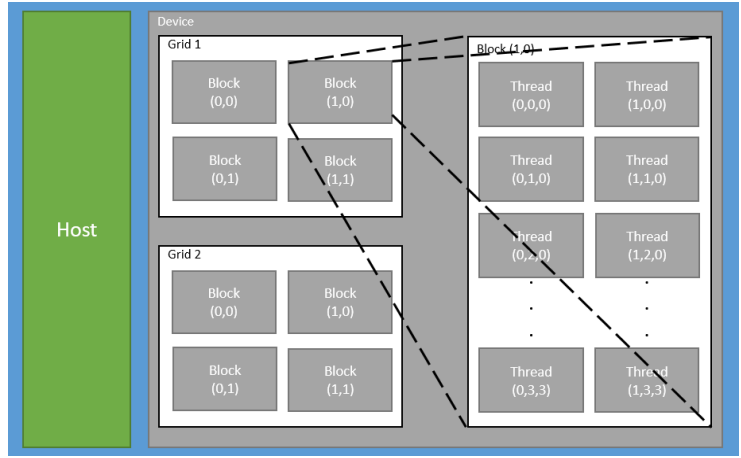

CUDA development model consists of combination CPU and GPU code. The CPU code is normally used for complex data structures and control flow while the GPU code is targeted for massively parallel operations such as SIMD operations (Single Instruction Multiple Data). The algorithms that are developed for GPU utilization consist of kernels that are designed to run on the GPU and control code which executes on CPU. The algorithms typically transfer large amounts of data to GPU and then run the CPU control code. The control code will execute kernels as appropriate and finally results will be moved back from GPU to CPU for display.

## 4. Parallel 3DPIFCM

In 3DPIFCM the goal was to parallelize bottleneck operations that take most time on CPU. Since the algorithm accepts an image as input and runs on every pixel on this image sequentially, the most costly operation is therefore the per pixel operation which we called the IFCM step function. Also, as can be examined in the original step function it's independent from other operations that were done in previous steps. As a result, it's a perfect candidate for penalization to be executed on the GPU as a kernel function.

The step function runs a single iteration of segmentation with preset λ, ξ parameters as part of the iterative optimization algorithm. This function $f(\bar{x}_i(t))$ as shown in algorithm 1 is running through each voxel in the slice of a particular z axis image and recalculating the membership matrix and cluster centers including the cost function as shown in equation (1).

For each iteration of the step function equations 2,3,4,5,6,7,8,9,10 need to be executed in order to calculate equation (1) and get the new cost which will be evaluated against previous iterations of $J_m(U,\bar{c})$ until reaching convergence. Our implementation of 3DPIFCM on the GPU is utilizing heavy caching of pre calculated feature maps and neighborhood maps that are calculated once at the initial steps of the algorithm and passed to the GPU as matrices. This reduces considerable GPU time for each calculation of voxel per iteration. Next we examine the algorithm and caching of parameters.

First we look at the new equations we introduced 5, 6, 7, 8, 10 for possible caching optimization. In equations 5, 7 we can see that $g_{ik}$, $q_{ik}^2$ and $W_r$ are matrices that are not modified between each step function $f(\bar{x}_i(t))$. Therefore it is possible to pre calculate them in the initial step on the CPU, transfer to the GPU and cache them for further iterations. Since the main membership and cluster centers equations (2), (3) are modified in each iteration of the step function they cannot be cached on the GPU. As a result they are also calculated on the GPU memory for efficient optimization and runtime. We show a flowchart of the 3DPIFCM algorithm on the GPU in Algorithm 2.

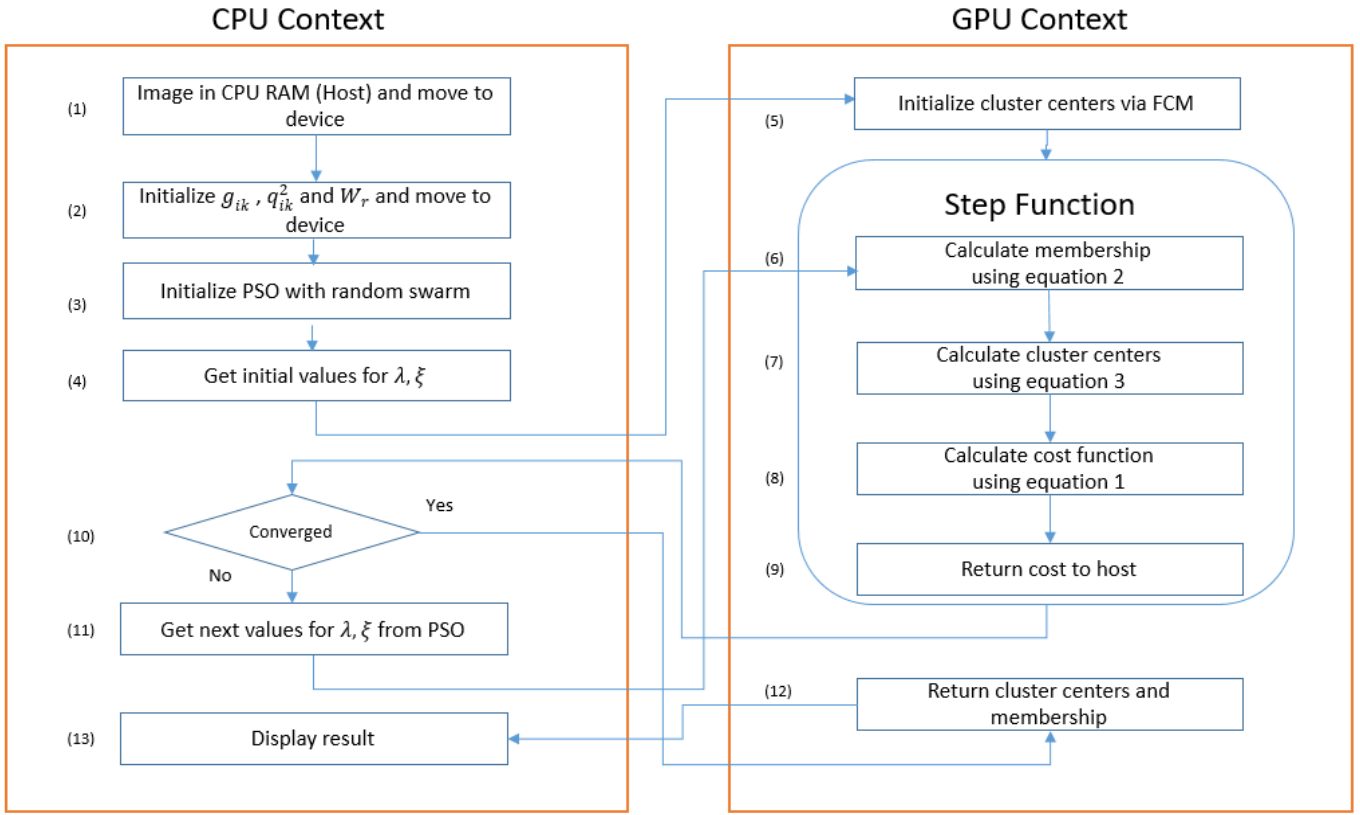

Algorithm 2: 3DPIFCM on GPU

Algorithm description:

<u>Input:</u>
   img3d – A 3D matrix of pixel intensities
   c – Number of clusters
   v – Depth parameter
   h – Exponential decay
   z – Slice in z axis to segment
   m – Fuzziness
   $\epsilon$ – Stop criteria
<u>Output:</u>
   Centers – the centroids of each cluster.
   U – membershipMatrix

<u>Flowchart:</u>
1. Prepare the image, normalize it and move to device memory (GPU).
   A normalize via min max so that each voxel value is between 0 and 1 across the entire 3D image. Consequentially, the results are more stable during the optimization of PSO and the cost function.
2. Initialize the parameters $g_{ik}$, $q_{ik}^2$ and $W_r$ from equations 5, 6, 7, 8, 10 and move the matrices to device memory. The cost of initialization of the parameters on the CPU is negligible to the entire runtime of the algorithm.
3. Initialize PSO swarm. According to the PSO algorithm as depicted in Algorithm 1 it's using velocity and position of the particle to optimize the cost function. This initialization ultimately creates random combinations of particles with different λ, ξ values ranging from 0 to 1.
4. Get initial values for λ, ξ from initial position and velocity of PSO swarm.
5. Execute FCM on device and get initial cluster centers and membership matrix (C, U respectively). As FCM runtime is negligible in comparison to rest of the algorithm we used an approach similar to [18] for FCM on GPU approach.
6. Calculate membership matrix on the GPU using equation 2. This calculation is running in parallel using thread per voxel approach as in [3].
7. Calculate cluster centers on GPU for values λ, ξ using equation 3 (which utilizes new equations 5-10). Here too we run thread per voxel as in [3].
8. Calculate Cost function using equation (1) on GPU.
9. Return the result of cost function to CPU for loop control.
10. Check if convergence of $\epsilon$ (algorithm 1) was reached in this iteration. If yes go to 12, otherwise go to 11.

11. Retrieve λ, ξ from PSO according to swarm velocity and position and go to step 6.
12. Get final clustering centers and membership result and move to host.
13. Display membership result on host.

## 5. Tools

The algorithm is implemented entirely with open source tools. We used python language with scientific optimization package numpy, scipy and JIT compiled extension Numba[10] with CUDA support. The Numba package allows us to compile functions to C code both for CPU and GPU usage. The nvvc annotations that are used within Numba are similar to the original C language compiler annotations that are present in C/C++ code. For example when JIT compiling a function to CPU usage we use an annotation such as @jit(nopython=True) and require limited python language to enable to execute it. In order to enable GPU execution we use @cuda.jit annotation before the function.

When using such an annotation before a function we ensure that this function becomes a GPU kernel function. In order to pass parameters to this function we first must transfer them to the GPU as a preprocessing step. Once this function is activated CUDA enables us to use parallelism of block and threads. In our implementation we assign a thread for each voxel that is being calculated against each cluster.

## 6. Experiments

### 6.1. Introduction

In order to test for parallel 3DPIFCM we wanted to test against various algorithms that were compared in [1]. The comparison matric is execution speed in seconds given that all other parameters are the same for each algorithm. All quality comparisons were done in the previous paper [1]. In order to ensure that the GPU version is comparative to the CPU version in quality few tests were conducted on various images and got similar segmentation results between the two.

The algorithms we compared against are shown in table 3. Each test was conducted several times and average speed was taken. Since the FCM version is doing no parameter optimization and noise reduction it was significantly faster than the rest of the algorithms including the GPU version. However, in the experiments we show that the GPU version can run in reasonable time whereas the other CPU generic variants significantly degrade in runtime as the images increase in size.

Table 3: algorithms comparison

| Algorithm Name | Description | Features | Hardware |
|---|---|---|---|
| **FCM[5]** | Standard Fuzzy C Means Clustering (implementation in Python) | 2D features of intensity difference between pixels only. This is original FCM which has no optimization parameters. | CPU only |
| **IFCMPSO[7]** | IFCM with Particle Swarm Optimization of parameters $\lambda$ and $\xi$. We use same parameters for PSO for IFCM-PSO, 3DPIFCM, 3DPIFCM-GPU. | 2D features, PSO optimization of $\lambda$ and $\xi$. This is original IFCM with PSO optimization which doesn't use $V, h, g$ parameters. | CPU only |
| **3DPIFCM[1]** | IFCM with Particle Swarm Optimization with 3D features and optimization parameters. This is our incarnation of the algorithm with 3D features for noise reduction. | 3D features, PSO optimization of $\lambda, \xi$. In addition, testing different configurations of $V, h, g$ parameters for performance tuning [1]. | CPU only |
| **3DPIFCM-GPU** | This is a new implementation of 3DPIFCM on a GPU. We use several techniques and move the image entirely to the GPU memory and perform parallel computation there. | 3D features, PSO optimization of $\lambda, \xi$. In addition, testing different configurations of $V, h, g$ parameters for performance tuning. | CPU and GPU combination. Sequential part on CPU and parallel on GPU. |

All the implementations of the algorithms were done in python with Numba[10] JIT compilation for the segmentation parts. The runtime is comparable to native C versions since we used heavy optimization of scientific libraries such as Numpy[12] and Numba which are native implementations. The GPU CUDA version was also JIT compiled using Numba directives on kernel functions. All algorithms we compared against including 3DPIFCM – CPU run on an 8 core CPU with 64 GB of memory and a TITAN X GPU with 3072 SP cores and 12GB of memory. The following sections describe the experiments.

*6.2. Synthetic Volume experiment*

*6.2.1. Goal*

The goal of this experiment was to show the relationship between size of an image and speed of execution between parallel version of 3DPIFCM and other IFCM variants including FCM.

*6.2.2. Method*

A synthetic dataset was used showed in [1]. The dataset is a cube with in a cube with ranging grayscale colors in 3D space. An image of this dataset is shown in figure 4.

Figure 4: Synthetic dataset. Cube of size 181x217x181 in grayscale.

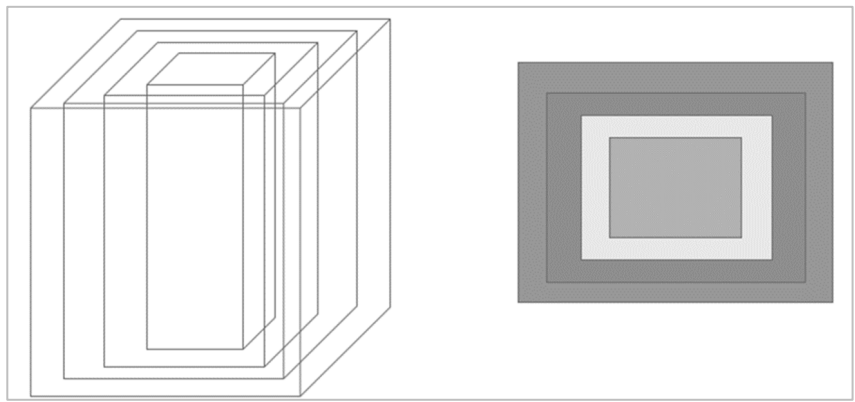

The size of the images in the original dataset is 181x217x181 voxels. In the experiments GPU speed was tested with ranging image sizes. Thus we recreated the dataset with varied sizes of 32x32, 55x55, 95x95, 165x165, 285x285, 493x493, 854x854. The images double in sizes to examine the speed with which all algorithms can handle the segmentation and optimization. Figure 5 shows the synthetic images.

Figure 5: synthetic images for execution speed comparison between various algorithms.

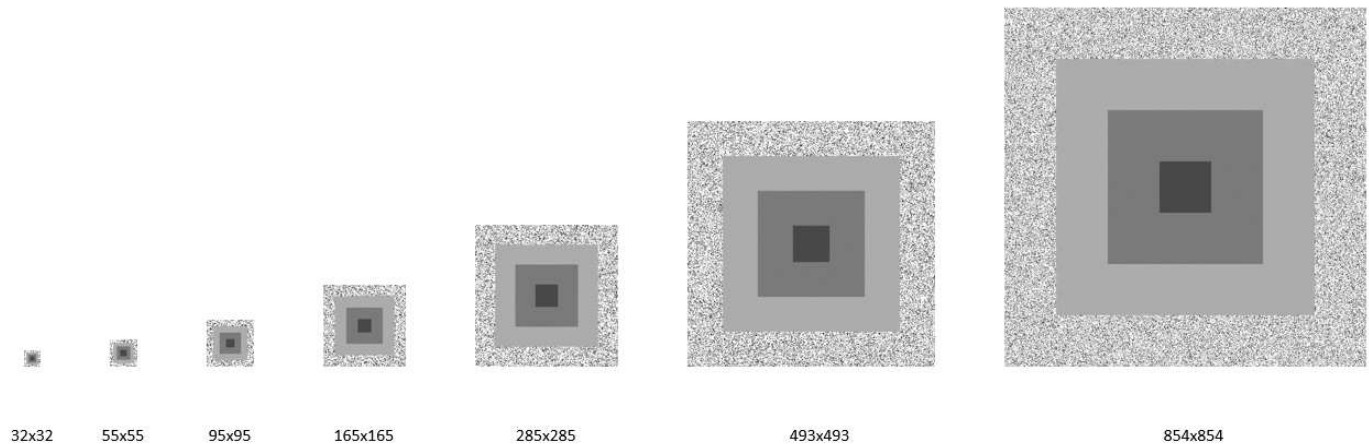

*6.2.3. Results*

A comparison of those different executions is shown in figure 6 and table 7.

Figure 6: Visual comparison in execution speed between all algorithms.

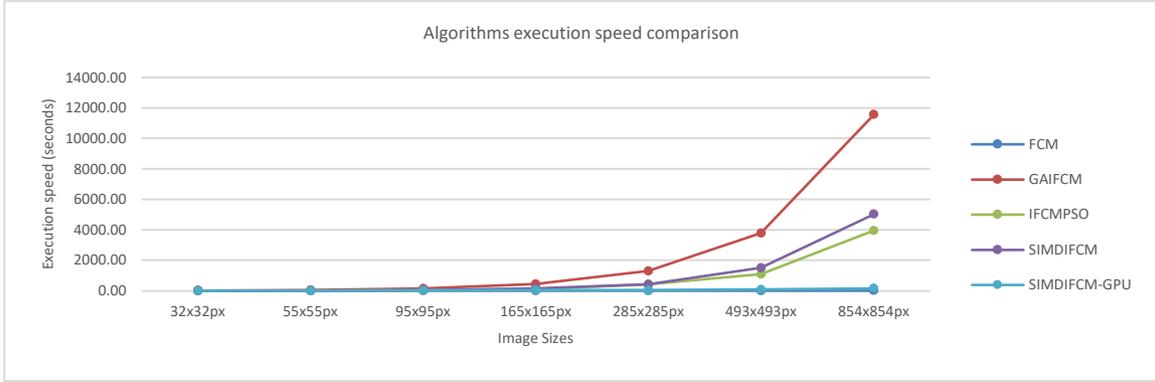

Table 7: execution time in seconds for all algorithms compared for synthetic data

| Size | FCM | GAIFCM | IFCMPSO | 3DPIFCM | 3DPIFCM-GPU |
|---|---|---|---|---|---|
| 32x32px | 0.02 | 15.95 | 4.62 | 6.66 | 7.44 |
| 55x55px | 0.05 | 49.20 | 11.24 | 12.75 | 10.80 |
| 95x95px | 0.15 | 146.39 | 40.53 | 46.86 | 17.27 |
| 165x165px | 0.53 | 442.48 | 99.58 | 142.55 | 41.57 |
| 285x285px | 1.79 | 1290.96 | 413.09 | 422.42 | 48.81 |
| 493x493px | 9.01 | 3794.71 | 1086.06 | 1501.91 | 89.47 |
| 854x854px | 24.56 | 11570.61 | 3958.05 | 5033.24 | 170.60 |

*6.2.4. Analysis of results*

As can be seen from Fig. 6 for small images there is a small overhead of moving the image to the GPU for parallel processing. As the images increase there is a significant benefit in running the GPU version. In fact we can see that for average image sizes of 285x285 there is an increase of 27x in speed compared to GAIFCM, 8.5x increase compared to IFCMPSO and 8.6x increase compared to the CPU version of 3DPIFCM. When we reach maximal size we see that the speed increase is 68x compared to GAIFCM, 23x for IFCMPSO and 29x for 3DPIFCM in CPU version. The advantage in these results is that as images in medical modalities become larger in sizes due to modern equipment and accuracy of devices parallel algorithms such as 3DPIFCM will start gaining more traction and implementation in clinical settings will be more frequent. The results show that 3DPIFCM is good candidate for clinical settings because of balance between speed and accuracy.

As can be seen in table 7 the FCM algorithm outperforms all IFCM variants including parallel 3DPIFCM in execution speed. However, as we showed in [1] it begins to degrade in accuracy as noise is introduced into the image. In order to adequately show a tradeoff between speed and quality in noisy images we create a cost function (11) to establish benefit of 3DPIFCM across different image sizes. The cost function trades off incorrect segmentation (incS) and execution speed in seconds (S).

$$J(\alpha) = \frac{1}{k}\sum_{i=1}^{k} \alpha\left(\frac{incS_i - \min(incS)_i}{\max(incS)_i - \min(incS)_i}\right) + (1-\alpha)\left(\frac{S_i - \min(S)_i}{\max(S)_i - \min(S)_i}\right) \quad (11)$$

We introduce a $\alpha$ variable to account for weight of incS [1] in regards to final cost. If $\alpha$ is close to 1 than most weight is contributed to quality and less for speed. If $\alpha$ is closer to 0 most weight contributed to speed. The variable k is the number of different image sizes, in our experiment k=7. Within the parenthesis of each expression we perform min max normalization of speed and incS to scale between 0 and 1. In figure 8 we show the cost function for different image sizes by averaging 3-9% noise levels. In this figure we set the $\alpha$ value to 0.7 to give the more weight to quality and less to speed.

Figure 8: Cost function $J(\alpha)$ for different image sizes with $\alpha = 0.7$.

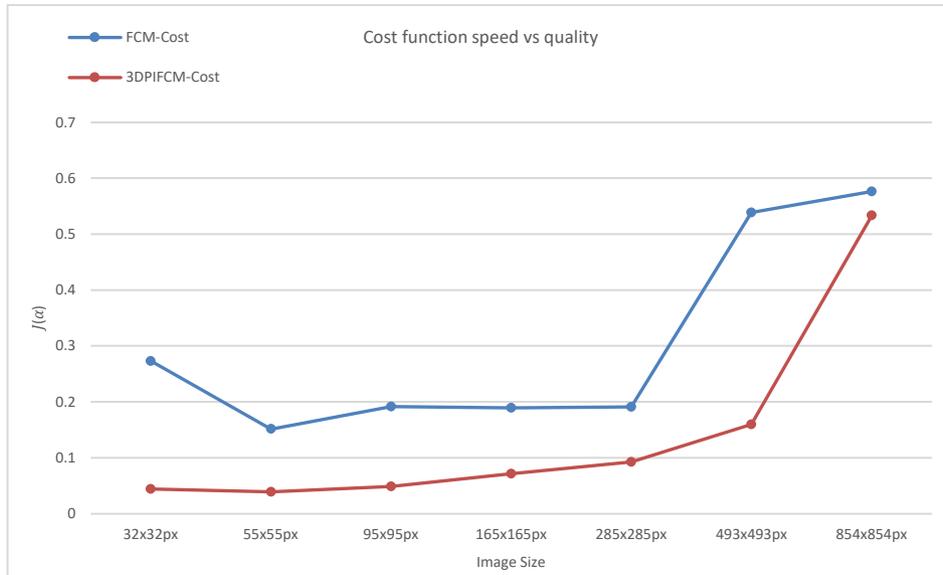

It can be seen that there is a constant margin in cost between FCM and 3DPIFCM in all noise levels on average. We also examine the cost function as a function of different values of $\alpha$ in figure 9. We can see that the cost of FCM is increasing above 3DPIFCM beyond a value of 0.45 and trending up while the cost of 3DPIFCM is trending down as $\alpha$ increases.

Figure 9: Average Cost function $J(\alpha)$ as a function of $\alpha$ for k different images.
Average Gaussian noises between 3% to 9%

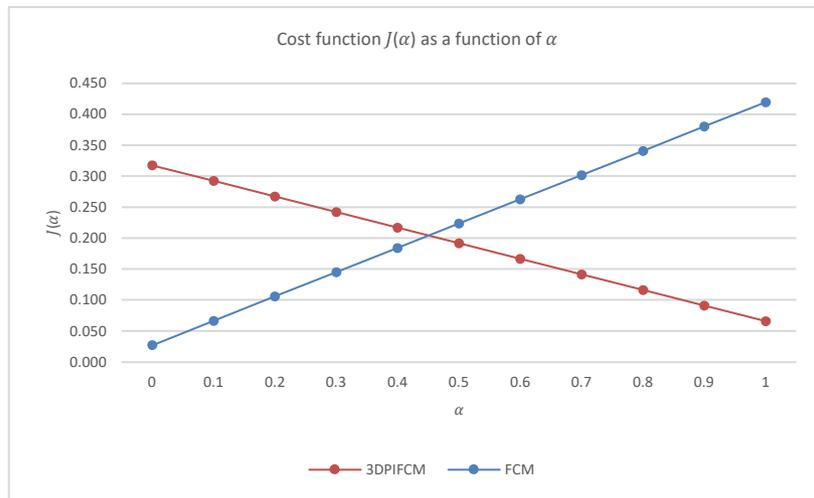

*6.2.5. Conclusion*

From experiments above it can be shown that 3DPIFCM parallel implementation is faster than all other IFCM variants when the image size is above 55x55 pixels. FCM outperforms parallel 3DPIFCM in speed but fails in quality when using noisy images. A tradeoff function was shown to compare between the two algorithms in terms of quality and speed of execution. We tested different levels of noise ranging between 3% to 9% Gaussian noise and averaged the results across levels. Above 9% noise the image loses information as we showed in [1]. Blow 3% there is no advantage to IFCM or it's variants because FCM performs good enough. The tradeoff function shows a clear advantage to 3DPIFCM for noisy images even when $\alpha$ is small and tends towards speed. This is primarily because the speed benefit of FCM is much less than the quality benefit of 3DPIFCM given the same conditions in noisy images. We can see a clear trend in ($\alpha$) with advantage to 3DIFCM as the quality component grows.

### 6.3. Brainweb Volume experiment

*6.3.1. Goal*

The goal of this experiment was to show realistic conditions by using a simulated MRI brain volume from Brainweb. Because real data such as Brain MRI scan contains many different gray levels and inconsistencies within colors it may present different results from a simulated image.

*6.3.2. Method*

Standard Brainweb [6] volume was used in this experiment. It was the same data used in [1]. The size of the 3D images there are 181x217 on the axial z axis. We show the Brainweb data in figure 10.

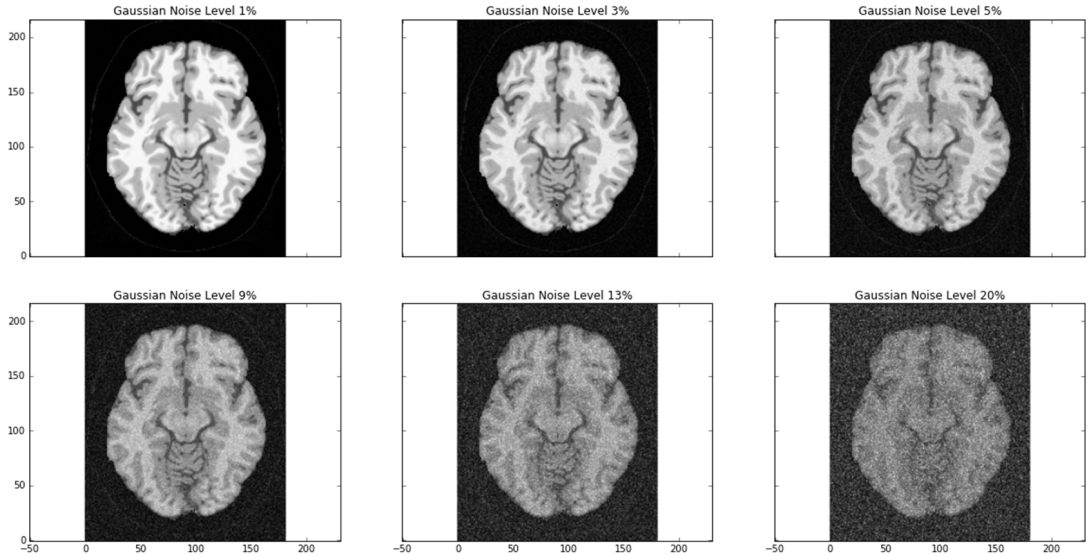

Figure 10: Brainweb data with Gaussian noise level from 1% to 20%.

*6.3.3. Results*

We can see the execution results from Brainweb in figure 11.

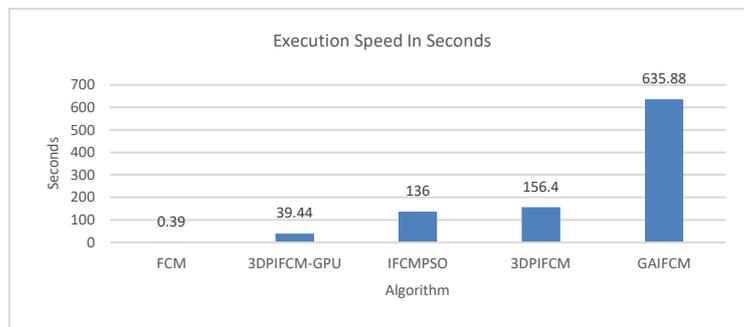

Figure 11: Comparison of algorithms in Brainweb slice 100 with 7% Gaussian noise.

*6.3.4. Analysis of results*

As we can see, the FCM version is performing considerably faster even when executing on a CPU with JIT compilation enabled. However, as we showed in [1] there is a degradation in segmentation quality for FCM when noise levels increase above 1%. The increase in quality of segmentation between FCM and 3DPIFCM on the GPU is 4.6x which is considerable when taking into account the medical implications of incorrect segmentation.

We can clearly see that the GPU version outperforms the CPU version by a factor of 4. When we compare to IFCMPSO on CPU version we get a 3.5x increase in speed and 16x increase compared to GAIFCM for the same volume. Since the speedup is dependent on specific hardware setup and GPU we would like to give a theoretical foundation to the potential benefits of parallel 3DPIFCM. We use a modified Amdahl's Law [4] to test the benefits because our algorithm has both sequential and parallel parts.

$$S_{latency}(s) = \frac{1}{(1-p) + \frac{kp}{jN}} \quad (12)$$

This formula gives the potential benefit of a parallel version of an algorithm given the following variables:

p – The proportion of execution time that the part benefiting from improved resources originally occupied.
N – Number of processors.
k – The ratio between CPU clock speed and GPU clock speed per processor.
j – The ratio of data transfer between CPU to GPU.

In order to show the benefits several assumptions were made regarding the parameters of this formula.
First, the ratio k is set to 3. This assumption is made by testing current modern GPU's which have roughly one third the clock speed of a CPU. Second, we set p to 0.99. p is the proportion of execution time that the part we make parallel uses. In case of 3DPIFCM 99% reflects that most time is spent on the kernel part. In addition, we set j which is the ratio between algorithm running time and data transfer to 1/50. In figure 12 we show the potential benefits of speedup as N grows. This means that we increase the number of GPU SP cores.

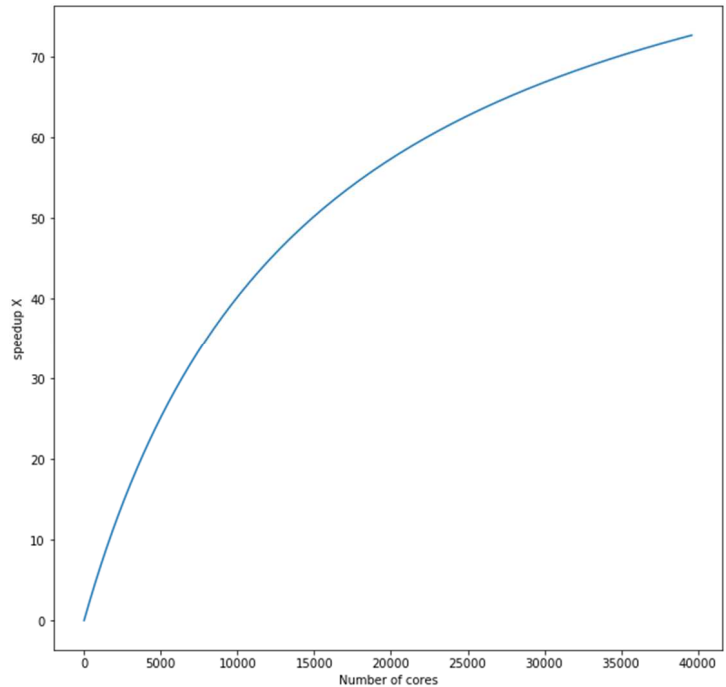

Figure 12: Speedup of 3DPIFCM as a function of number of cores

*6.3.5. Conclusion*

In this experiment it was shown that 3DPIFCM can perform well in real world conditions against known IFCM variants while increasing speed. Also, it was shown that the algorithm can be used in clinical settings. In real MRI images the Gaussian noise factor normally ranging between 1-10% [6]. When executing 3DPIFCM in parallel mode on the GPU we saw a constant ratio in speed increase regardless of the noise component in the image. This can be shown well both in Fig. 6 and Fig. 11.

# 7. Limitations

The limitations of our study were:

1. Using a specific hardware and software setting to compare speed. This can be difficult to generalize exact speedups in different environments with different constraints.
2. We used a limited parallel environment by using a single GPU. Theoretically many GPUs can be used if distributing the computation correctly. This could lead to additional potential speedups.

Overall our study represents a general approach to clustering given noisy artifacts in medical images. The limitations presented above could all be mitigated by change of hardware and software, and lastly additional computational resources with many GPUs as setup.

# 8. Conclusions

In this paper we presented a new parallel algorithm which is based on 3DPIFCM [1]. The algorithm can segment noisy medical images for different modalities in a parallel environment namely CUDA GPUs. The algorithm allows for physicians and medical staff to use fuzzy clustering for noisy images in a clinical setting without compromising time as compared with other sequential variants.

We showed that although the GPU version is comparative in quality it's up to 68x faster than the slowest IFCM variant. For average sized images of 285x285 it's 15x faster than other IFCM variants on average including the same algorithm's implementation on a CPU. When we compared the algorithms on a Brainweb T1 volume of size 181x217, axial view, we reached an increase in 4.6x in quality compared to FCM and speedup of up to 16x compared to GAIFCM and 3.5x compared to IFCMPSO.

In addition, we gave theoretical foundation using Amdahl's Law with certain assumptions suggesting a much bigger speedup given more SP cores and faster GPU. In this setting it is possible to reach 100x speedup by using 15000 SP cores in the future instead of 3500 as of writing this paper.

Our main contribution in this paper is a parallel version of 3DPFICM which performs well in clinical settings and achieves state of the art quality segmentation in noisy images. We introduced a new cost function to compare between two algorithms by factoring noise and segmentation quality. The 3DPIFCM algorithm has a clear advantage over FCM when we compare against speed and quality in noisy images. This algorithm opens a new door for clinical parallel algorithms for image segmentation that utilize modern computational environments.

# 9. Future steps

Some further steps for our research include a fully native implementation of 3DPIFCM on either C or C++ which may increase execution performance further. In addition, we would like to utilize shared memory as shown in [3] for better utilization of the GPU internals. The shared memory as shown in [3] is 100 times faster than global device memory and can significantly increase the step function's performance further. Also, we would like to test on different images modalities such as CT and X-Ray of different sizes and of different organs. This would validate our approach for noisy images segmentation and promote a clinical settings experiment.